\documentclass{article} 
\usepackage{iclr2027_conference,times}

\usepackage{amsmath,amsfonts,bm}

\def\eqref#1{equation~\ref{#1}}

\def\1{\bm{1}}

\DeclareMathAlphabet{\mathsfit}{\encodingdefault}{\sfdefault}{m}{sl}
\SetMathAlphabet{\mathsfit}{bold}{\encodingdefault}{\sfdefault}{bx}{n}

\usepackage{hyperref}
\usepackage{url}
\usepackage{graphicx}
\usepackage{booktabs}

\title{Hierarchical Diffusion for \mbox{Camera-controlled} Autoregressive Video Generation}

\author{
Kaihua Chen$^{1,2}$\thanks{Work done while at Carnegie Mellon University.} \quad
Yiming Gong$^{1}$ \quad
Deva Ramanan$^{1}$ \quad
Tarasha Khurana$^{1*}$ \\
$^{1}$Carnegie Mellon University, Pittsburgh, PA, USA \\
$^{2}$University of Oxford, Oxford, UK
}

\iclrfinalcopy 
\begin{document}

\maketitle

\begin{abstract}
We address the task of long-video, camera-controlled generation from a single image. The dominant autoregressive paradigm suffers from two key limitations: (1) error accumulation over time, as distant frames are conditioned on previously generated outputs rather than ground truth, leading to drift; and (2) ambiguity in how to incorporate camera control. Standard solutions to both—such as self-forcing and conditioning on explicit 6-DOF poses—either incur significant computational overhead or produce drifting scene geometries over time. 
We propose a simple and lightweight framework that combines autoregressive generation (WanAR) with hierarchical long-context denoising (FixAR).
To enforce scene geometry remains consistent with
input cameras, we draw from feed-forward novel-view synthesis and reformulate video generation as a warping-and-inpainting problem. To mitigate drift, we introduce a hierarchical denoising strategy in which different stages condition on different temporal histories. Specifically, we first generate frames conditioned on the immediate past, and then refine them using a second denoiser conditioned on a temporally subsampled long-range history to enforce global consistency. This fine-to-coarse scheme enables each generated frame to effectively condition on the full temporal context without incurring a prohibitive cost.
Experiments on RealEstate10k and DL3DV demonstrate that WanAR+FixAR outperforms prior methods for camera-controlled long-video generation, producing results that are both temporally stable and consistent with 
input camera trajectories.
\end{abstract}

\section{Introduction}

Recent advances in video diffusion models~\cite{blattmann2023stablevideodiffusion, yang2024cogvideox, wan2025wan, hacochen2024ltxvideo} have enabled photorealistic video generation with increasing levels of control. In this work, we study the problem of long-video generation from a single image given future camera trajectories. This setting is particularly appealing for applications such as virtual scene exploration and content creation, where controllable viewpoint changes must remain 3D consistent over extended time horizons. However, generating long videos that faithfully follow camera motion while maintaining spatial and temporal coherence remains a significant challenge.

\noindent\textbf{Why is this hard?} Existing approaches predominantly rely on autoregressive generation, where future frames are conditioned on previously generated outputs over a sequence of rollouts. While this is effective for short sequences, such methods accumulate errors over time, leading to drift in long videos. A second challenge lies in how camera motion is incorporated. One widely adopted approach is to condition on explicit 6-DOF pose representations, such as Plücker coordinates or ray maps. One subtle challenge is that many flagship benchmarks (like RealEstate10K~\cite{zhou2018re10k}) provide cameras only up-to-scale, making camera-conditioned video generation somewhat undefined - e.g., the world looks different when moving forward 10cm versus 10m.

\noindent\textbf{Status quo.} Prior work has attempted to address these challenges along two main directions. To mitigate autoregressive drift, several methods adopt self-forcing strategies~\cite{huang2025selfforcing}, where models are periodically trained or evaluated on their own generated outputs. While effective in reducing error accumulation, these approaches significantly increase computational cost and are difficult to scale to long sequences; there is a linear increase in time and space for generating training data as rollouts can only be sequentially obtained. Most camera-controlled video generation methods appear to circumvent the up-to-scale ambiguity by overfitting to the scale of a particular dataset, making such camera-conditioned generation models unlikely to generalize.

\noindent\textbf{Our approach.} We address these challenges with a simple and lightweight framework that combines online reconstruction with hierarchical denoising. First, to respect the fixed but unknown scale of input camera trajectories, we draw inspiration from feed-forward novel-view synthesis and reformulate video generation as a warping-and-inpainting problem, aligning intermediate depth predictions to the scale of the ground-truth geometry (typically provided in the same scale as camera inputs, often obtained via COLMAP).
Second, to mitigate autoregressive drift, we propose a hierarchical denoising strategy in which different stages condition on different temporal histories. Concretely, we generate frames in a fine-to-coarse manner: an initial denoiser conditions on the immediate past to produce high-frequency details, while a second denoiser conditions on a temporally subsampled long-range history to enforce global consistency. This design allows each generated frame to efficiently incorporate information from the entire past without incurring the cost of conditioning on an exhaustive set of historical frames.

In summary, our contributions are as follows: (1) we identify scale inconsistency as a key failure mode of pose-conditioned video generation and propose a warping-based formulation that enforces scale consistency between the scene and camera trajectory; (2) we introduce a hierarchical denoising scheme that reduces long-horizon drift by conditioning on multi-scale temporal context; and (3) we present a simple combination of models, WanAR and FixAR, that achieves state-of-the-art performance on RealEstate10k and DL3DV for camera-controlled long-video generation, producing videos that remain consistent with the scale of input camera trajectories. 


\begin{figure}
    \centering
    \includegraphics[width=\linewidth, height=5cm]{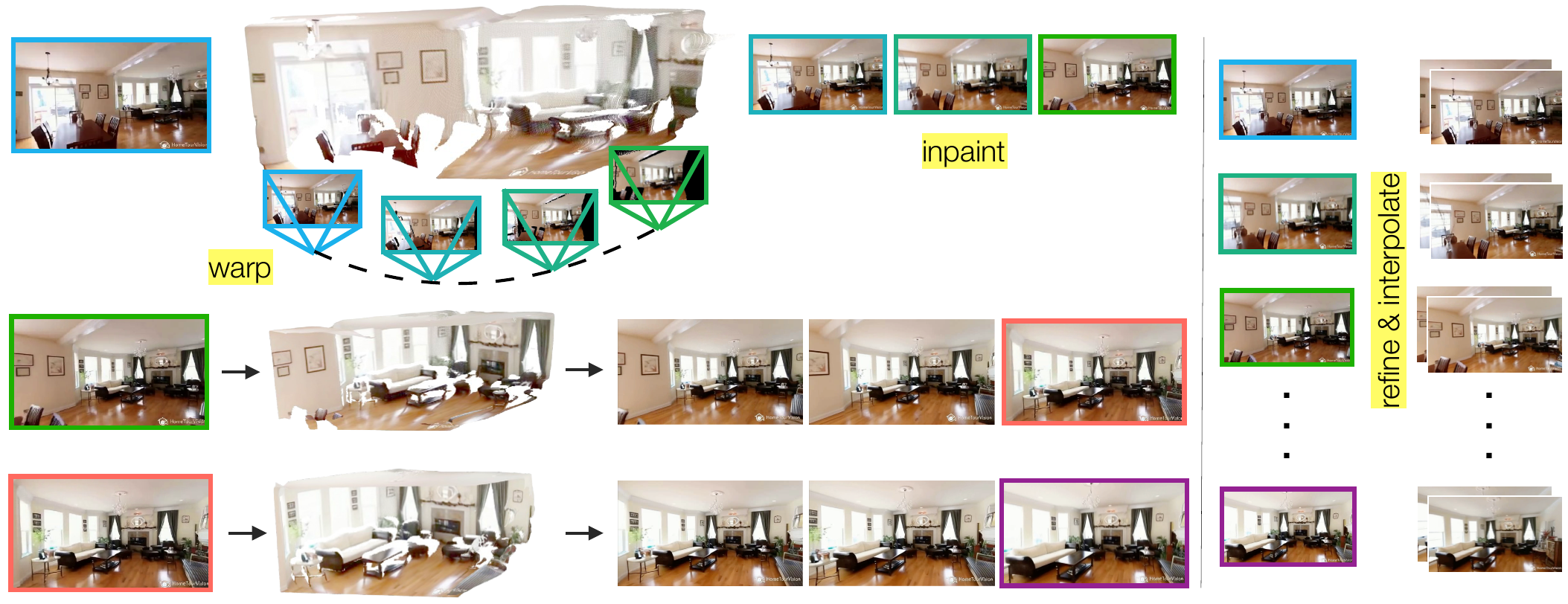}
    \caption{In this work, we consider the problem of camera-controlled video generation. Existing approaches condition on cameras represented as ray encodings or Plucker coordinates, but these learn poses up to a global scale factor. We draw from literature in feed-forward novel-view synthesis that warp given views into future poses and reformulate video generation as video inpainting (\textbf{left}). While this allows for precise camera control, minute-long autoregressive video generation in this setup still suffers from drift and error accumulation over time. We propose a hierarchical denoising scheme (\textbf{right}) that looks at multiple rollouts from the inpainting stage and corrects for long horizon artifacts. }
    \label{fig:teaser}
\end{figure}

\section{Related Work}

\noindent\textbf{Autoregressive long video generation.}
Modern video diffusion models~\cite{ho2022videodiffusion,blattmann2023stablevideodiffusion,hacochen2024ltxvideo,yang2024cogvideox,wan2025wan,chen2025diffvas} can generate photorealistic videos, but are typically limited to short temporal horizons. To extend generation length, autoregressive formulations are widely adopted to roll out longer sequences~\cite{chen2025skyreelsv2, teng2025magi1, deng2024autoregressive, jin2024pyramidal, chen2024diffusionforcing, yin2025causvid, song2025dfot, huang2025selfforcing, cui2025selfforcingpp}. However, a fundamental challenge in autoregressive video diffusion is \textit{drift}, which progressively degrades visual quality over long rollouts. Recent approaches attempt to mitigate this issue by either exposing models to drifted distributions during training~\cite{chen2024diffusionforcing, song2025dfot, huang2025selfforcing, zhu2026causalforcing, liu2025rollingforcing}, or by modifying attention mechanisms and introducing memory compression modules to reference earlier, cleaner context during inference~\cite{zhang2025framecontext, cai2025mixturecontexts, zhang2025framepreservation, yang2025longlive, li2026rollingsink}. While these methods improve long-range consistency, they still suffer from limitations. For example, diffusion-forcing~\cite{chen2024diffusionforcing, song2025dfot} relies on approximated drift distributions and exhibits a train–test drift distribution mismatch, while self-forcing~\cite{huang2025selfforcing} incurs increased training cost due to multi-step rollouts at every training iteration. In this work, we propose to decouple autoregressive generation (WanAR) and drift correction (FixAR) via a hierarchical denoising framework, where FixAR is trained on precomputed data pairs with real drift.

\noindent\textbf{Camera-conditioned video world model.} 
By introducing action controllability into long video generation and making the underlying world behind visual content explorable, video world models have emerged and demonstrated strong performance~\cite{bruce2024genie, ball2025genie3, valevski2024diffusiongame, hyworld2026, hong2025relic, team2026lingbotworld}. Most video world models primarily used keyboard inputs as a simple form of action control~\cite{he2025matrixgame, guo2025mineworld, openoasis2025}, while many recent works ~\cite{yu2025contextmemory, xiao2025worldmem, kong2025worldwarp, li2025vmem, zong2020lyra2, wu2025videoworldmemory, gao2025longvie2} increasingly emphasizes camera motion as the control signal to enable more precise manipulation and support specific application scenarios. A natural question is how to incorporate camera information into video generative models. Existing approaches can be broadly categorized into three groups: (i) direct camera encoding, (ii) injecting 3D geometry through positional encoding fields~\cite{li2025prope, bai2025positionalfield}, and (iii) injecting 3D geometry via rendering in pixel space. Most existing methods~\cite{yu2025contextmemory, xiao2025worldmem, shen2026lyra, li2025vmem, team2026lingbotworld, hyworld2026, hong2025relic, he2025matrixgame, hong2025relic, yu2025contextmemory} fall into the first category, where normalized representations (e.g., Plücker rays or ray maps) are encoded into latent space to guide the generation process, allowing camera control to be learned from large-scale camera annotations. However, such latent-space conditioning often leads to imprecise control when reflected in pixel-space generation, and tends to generalize poorly to out-of-distribution camera trajectories. The second line of work~\cite{shen2026lyra, zhang2025UCPE} models 3D geometric transformations induced by camera motion through modifications to positional encoding. This approach achieves more accurate camera control, but the modified architecture introduces additional training overhead. In this work, we adopt the third line of work~\cite{wu2025videoworldmemory, kong2025worldwarp}, which directly models camera control by estimating 3D geometry using a metric depth estimator and warping it into future frames in an online fashion. This approach enables accurate camera control while naturally reformulating autoregressive generation as an inpainting task.

\noindent\textbf{Generative novel-view synthesis.} 
Current state-of-the-art novel-view synthesis (NVS) methods are largely based on generative diffusion models~\cite{gao2024cat3d, wu2025cat4d, watson2022nvsdiffusion, yu2025viewcrafter, yu2025trajectorycrafter, chen2025cognvs, bai2025recammaster, he2025cameractrlii, ren2025gen3c, li2026rerope, bai2025positionalfield, seo2024genwarp, zhao2024metricfromhuman}. They are closely related to camera-conditioned video world models, but their spatial and temporal horizons for novel-view generation remain limited. Early NVS approaches also widely adopted direct camera encoding in latent space~\cite{gao2024cat3d, wu2025cat4d}, but lacked precise camera control. More recent methods instead employ 3D geometry-based warping-and-inpainting formulations~\cite{yu2025viewcrafter, yu2025trajectorycrafter, chen2025cognvs, ren2025gen3c}, leading to improved visual quality and more accurate camera control.

In addition, there is a line of work in NVS that uses diffusion models to correct rendering artifacts caused by large viewpoint changes~\cite{wu2025difix3dplus, delutio2026artifixer}. This perspective is closely related to the motivation behind our FixAR. In our setting, we interpret autoregressive drift in long-horizon video generation as another form of accumulated artifact over time, which can be corrected using video diffusion.

\section{Method}
\label{sec:method}

\subsection{Preliminaries: Flow matching video diffusion}

We adopt the open-source DiT-based~\cite{peebles2023dit} video diffusion model Wan-I2V-14B~\cite{wan2025wan} as our base model. Unlike traditional diffusion models that rely on score matching and stochastic reverse diffusion, Wan is trained with rectified flow matching~\cite{esser2024sd3, liu2023rectifiedflow, lipman2023flowmatching}, which formulates generation as solving an ordinary differential equation (ODE) without stochastic terms. This formulation provides a stable and scalable training objective while preserving a likelihood-based interpretation.

Specifically, given a clean video latent $x_1$, a random noise latent $x_0 \sim \mathcal{N}(0, I)$, a conditioning signal $c$, and a timestep $t \in [0,1]$, flow matching defines the interpolated latent as
\[
x_t = (1-t)x_0 + t x_1 .
\]
The model is trained to predict the constant velocity field that transports noise $x_0$ to data $x_1$:
\[
\mathcal{L}
=
\left\|
f_{\theta}(x_t, c, t) - (x_1 - x_0)
\right\|_2^2 .
\]

\begin{figure}
    \centering
    \includegraphics[width=1\linewidth]{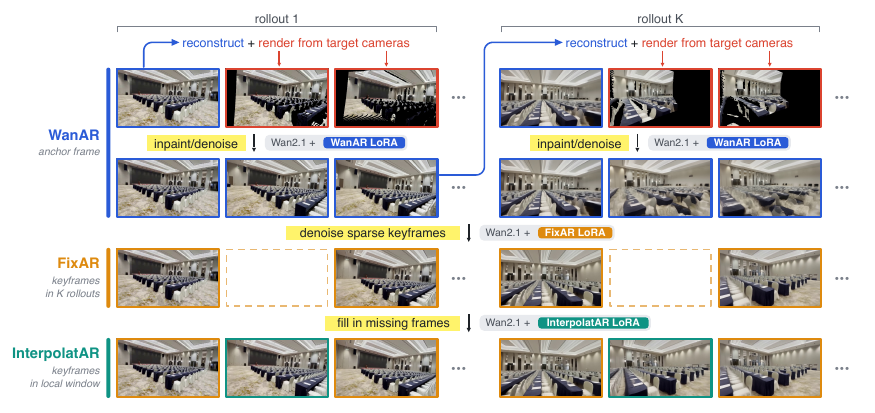}
    \caption{\textbf{Overview of WanAR, FixAR, and InterpolatAR.} We build upon WAN~\cite{wan2025wan}, a video diffusion model, and adapt it for camera-controlled autoregressive video generation in static 3D scenes. All three stages share the same Wan2.1 backbone weights, with a different stage-specific LoRA checkpoint loaded for each stage. At each rollout, we estimate depth using MoGe-2~\cite{wang2025moge2} on the input frame (or the last generated frame from the previous rollout), and warp it into future camera poses to provide geometry-aware conditioning for WanAR (\textbf{top}). While WanAR can generate plausible long videos autoregressively, visual drift and artifacts gradually accumulate over multiple rollouts. To mitigate this issue, we introduce a hierarchical denoising framework with a second-stage denoiser, FixAR (\textbf{middle}), which refines temporally subsampled keyframes generated across different rollouts using a broader temporal context. Finally, since FixAR only operates on sparse keyframes, we apply InterpolatAR to synthesize intermediate frames between adjacent refined keyframes (\textbf{bottom}). The entire three-stage pipeline can itself be rolled out autoregressively to further extend the video. This hierarchical fine-to-coarse formulation enables both accurate camera control and temporally stable long-horizon video generation.}
    \label{fig:method}
\end{figure}

\subsection{WanAR: Camera control for autoregressive video generation}

We consider the problem of camera-conditioned long video generation: given an input image $I_0$ and a target camera trajectory $\{C_t\}_{t=0}^{T}$, the goal is to generate future observations $\{I_t\}_{t=1}^{T}$ that follow the specified camera motion, where $T$ can be large. General-purpose video diffusion models such as Wan-I2V-14B cannot be directly applied to this setting, because they (i) do not explicitly support camera conditioning, and (ii) are limited to a short generation horizon $L$, where typically $L \ll T$. To address this, we build \textbf{WanAR} on top of Wan. WanAR achieves camera control by propagating explicit 3D geometry (depth and camera), and extends generation length through an autoregressive formulation.

Given the initial observation $I_0$, we first estimate its metric depth $D_0$ using the monocular depth estimator MoGe-2~\cite{wang2025moge, wang2025moge2}. Since the corresponding camera $C_0$ is known, this gives us sufficient 3D geometry for the first frame. We then propagate this information to future viewpoints within the base model horizon by forward warping:
\[
(\tilde I_t, M_t)=\mathcal{W}(I_0, D_0, C_0, C_t), \qquad t=1,\dots,L-1,
\]
where $\tilde I_t$ denotes the warped partial observation at time $t$, $M_t$ is the corresponding inpainting mask, and $\mathcal{W}$ is the warping function.

Although the warped frames contain only partial RGB observations, they already provide accurate camera-conditioned spatial guidance. As a result, camera-conditioned video generation is reformulated as an inpainting problem that completes these partial observations:
\[
\{\hat I_t\}_{t=1}^{L-1}
=
f_{\theta}\!\left(I_0,\{(\tilde I_t, M_t)\}_{t=1}^{L-1}\right),
\]
where $\hat I_t$ denotes the generated full frame. To adapt Wan-I2V to this task, we modify its conditioning mechanism. In the original I2V model, the future frames and masks are represented by zero-padded latents in VAE space. In contrast, we replace this zero padding with the actual latent encodings of the warped frames and their corresponding inpainting masks.

So far, this formulation enables camera control only within the base diffusion horizon. To extend it beyond length $L$, we apply the same inpainting process autoregressively. At rollout $k$, we take the latest anchor frame $\hat I_{k(L-1)}$, estimate its depth online using MoGe-2~\cite{wang2025moge, wang2025moge2}, and warp it to the next chunk of target cameras. Formally, for
\[
k=0,\dots,\left\lceil \frac{T}{L-1}\right\rceil-1,
\]
we generate
\[
\{\hat I_t\}_{t=k(L-1)+1}^{(k+1)(L-1)}
=
f_{\theta}\!\left(
\hat I_{k(L-1)},
\{(\tilde I_t, M_t)\}_{t=k(L-1)+1}^{(k+1)(L-1)}
\right).
\]
In this way, WanAR produces a long video sequence that follows the entire target camera trajectory.

\subsection{FixAR: Hierarchical denoising for mitigating drift}

The autoregressive formulation of WanAR enables long-horizon generation, but it also introduces drift: except for the first rollout, each new chunk is conditioned on previously synthesized frames, so errors accumulate over time and gradually degrade visual quality. To mitigate this effect, we introduce \textbf{FixAR}, a second-stage fixing module that is applied after every few WanAR rollouts.

Specifically, after every $K$ rollouts, we uniformly sample a sparse set of keyframes from the generated sequence and feed them into FixAR for content correction. Let $\{\hat I_{\tau_j}\}_{j=0}^{L-1}$ denote the selected sparse keyframes. FixAR outputs the corrected frames
\[
\{I_{\tau_j}^{\star}\}_{j=0}^{L-1}
=
g_{\psi}\!\left(\{\hat I_{\tau_j}\}_{j=0}^{L-1}\right),
\]
where $g_{\psi}$ denotes the FixAR model and $I_{\tau_j}^{\star}$ are the corrected keyframes.

To train FixAR on realistic drift patterns, we construct its training set by following the same WanAR rollout process and the same sparse keyframe sampling strategy used at inference time. Unlike self-forcing approaches~\cite{huang2025selfforcing}, FixAR training does not require multi-step inference during each training iteration; instead, it uses precomputed input-target pairs, which substantially reduces computational overhead. Once trained, FixAR can be applied either globally to an entire generated sequence or autoregressively at a low frequency.

FixAR mitigates drift for two reasons. First, due to its hierarchical design, it can directly reference earlier frames with higher visual quality, or even ground-truth frames during training, allowing appearance and style to be propagated over a longer horizon. Second, because it is trained on realistic drift distributions, it learns to identify artifact-prone regions and correct them accordingly.

\noindent\textbf{Recovering the original video framerate.}
Because of the limited base model horizon and for computational efficiency, FixAR operates only on sparse keyframes. To recover the full video sequence at original framerate, we train an additional video diffusion model, \textbf{InterpolatAR}, to interpolate between neighboring corrected keyframes. This is a substantially easier task, since adjacent corrected keyframes are already visually and temporally close, and little hallucination is required. Importantly, InterpolatAR does not introduce additional drift, because it only fills in intermediate content rather than performing autoregressive extrapolation as in WanAR.

\section{Empirical Analysis}

\begin{figure}
    \centering
    \includegraphics[width=\linewidth]{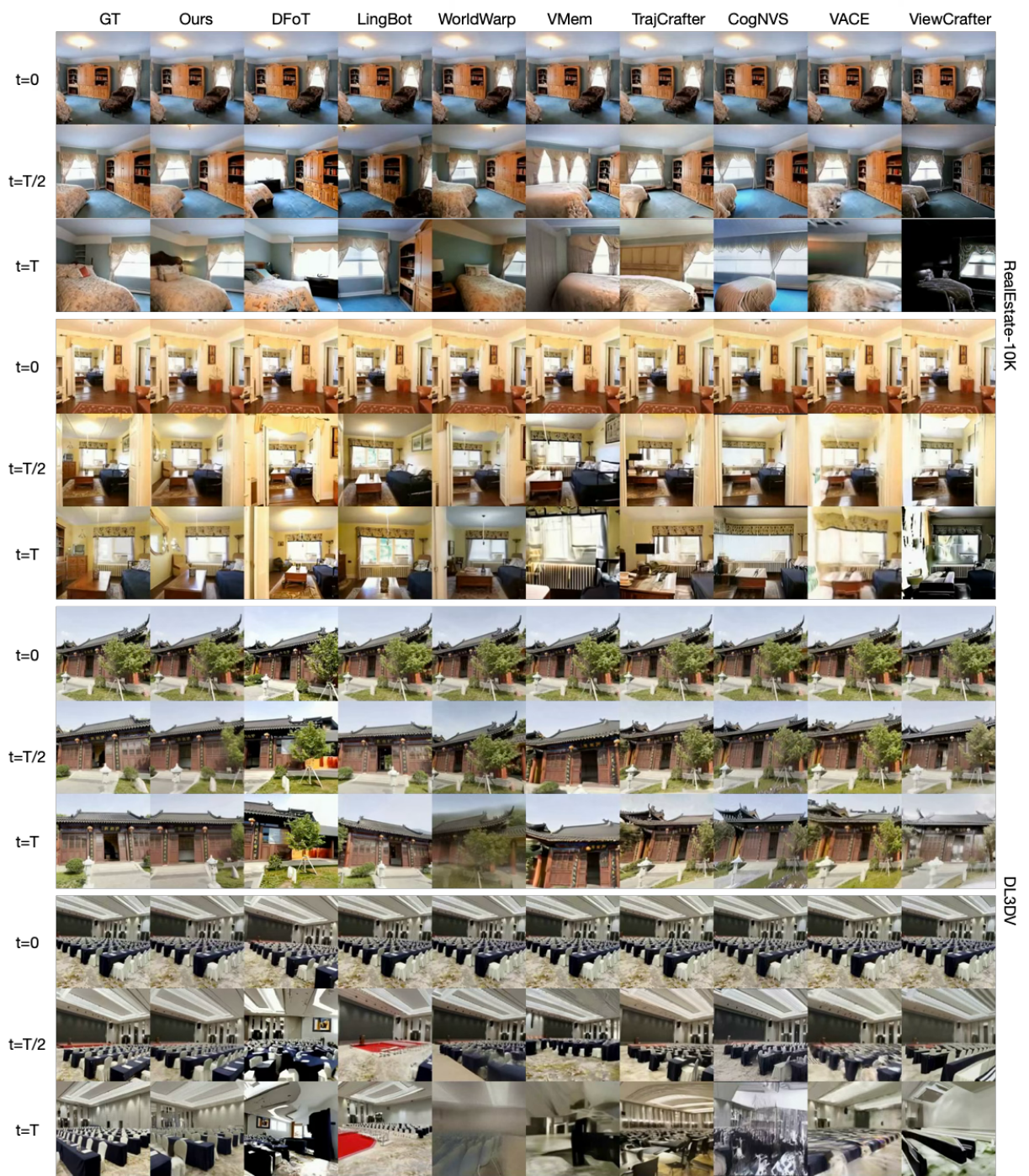}
    \caption{\textbf{Qualitative comparison on RealEstate10k and DL3DV.} We show qualitative results on two test scene from RealEstate10k (top two examples) and DL3DV (bottom two examples) across a range of video generation, and novel-view synthesis baselines. We see that proposed WanAR + FixAR combination of models ``renders" the scene using scale-consistent camera poses (as compared to ray encoding in DFoT or Plucker conditioning in LingBot), and show that hierarchical denoising mitigates drift over time which some autoregressive baselines like VMem struggle with. We also note that TrajectoryCrafter, a method designed for novel-view synthesis (in short sequences) is also surprisingly strong at the task of camera-controlled minute-long video generation.}
    \label{fig:qual_comp}
\end{figure}

\subsection{Experimental setup}
\label{sec:setup}

\begin{table}[t]
\centering
\scriptsize
\setlength{\tabcolsep}{3pt}
\resizebox{0.85\textwidth}{!}{
\begin{tabular}{lccccccc}
\toprule
\textbf{Method} 
& \textbf{PSNR} $\uparrow$ 
& \textbf{SSIM} $\uparrow$ 
& \textbf{LPIPS} $\downarrow$ 
& \textbf{FID} $\downarrow$ 
& \textbf{KID} $\downarrow$ 
& \textbf{FVD} $\downarrow$ 
& \textbf{mPSNR} $\uparrow$ \\
\midrule

ViewCrafter~\cite{yu2025viewcrafter} 
& 12.45 & 0.361 & 0.499 
& 67.36 & 0.0267 & 595.25 & 13.44 \\

TrajCrafter~\cite{yu2025trajectorycrafter} 
& 14.35 & 0.428 & 0.410 
& 30.54 & 0.0042 & 347.75 & 15.54 \\

CogNVS~\cite{chen2025cognvs}      
& 13.22 & 0.379 & 0.474 
& 48.58 & 0.0116 & 457.35 & 14.39 \\

VACE~\cite{jiang2025vace}        
& 15.23 & 0.487 & 0.426 
& 49.62 & 0.0119 & 393.30 & 16.89 \\

\midrule

Lingbot-World~\cite{team2026lingbotworld} 
& 12.90 & 0.362 & 0.513 
& 25.46 & \textbf{0.0012} & 398.31 & 13.36 \\

VMem~\cite{li2025vmem}          
& 12.13 & 0.336 & 0.583 
& 49.12 & 0.0099 & 465.13 & 12.46 \\

DFoT~\cite{song2025dfot}         
& 13.95 & 0.423 & 0.422 
& 38.28 & 0.0084 & 387.21 & 15.17 \\

WorldWarp~\cite{kong2025worldwarp}    
& 13.81 & 0.425 & 0.475 
& 38.59 & 0.0042 & 451.01 & 14.55 \\

\midrule

\textbf{WanAR + FixAR} 
& \textbf{16.69} & \textbf{0.525} & \textbf{0.341} 
& \textbf{23.16} & 0.0018 & \textbf{282.59} & \textbf{18.31} \\

\bottomrule
\end{tabular}}
\vspace{0.5em}
\caption{\textbf{Comparison to state-of-the-art on RealEstate10K~\cite{zhou2018re10k}.} We compare our hierarchical diffusion strategy to state-of-the-art approaches for long video generation and novel-view synthesis of static scenes. Our first finding is that a combination of WanAR + FixAR beats all prior art on RealEstate10K, which we attribute to better camera control in WanAR, and better drift mitigation from FixAR. One surprising finding is that novel-view synthesis approaches like TrajectoryCrafter in fact perform better at this task than many long video generation works. With mPSNR, we show better retention of visible regions particularly driven by precise camera control.} 
\label{tab:sota_re10k}
\end{table}

\begin{table}[t]
\centering
\scriptsize
\setlength{\tabcolsep}{3pt}
\resizebox{0.85\textwidth}{!}{
\begin{tabular}{lccccccc}
\toprule
\textbf{Method} 
& \textbf{PSNR} $\uparrow$ 
& \textbf{SSIM} $\uparrow$ 
& \textbf{LPIPS} $\downarrow$ 
& \textbf{FID} $\downarrow$ 
& \textbf{KID} $\downarrow$ 
& \textbf{FVD} $\downarrow$ 
& \textbf{mPSNR} $\uparrow$ \\
\midrule

ViewCrafter~\cite{yu2025viewcrafter} 
& 13.55 & 0.307  & 0.459
& 74.76 & 0.0114 & 847.25 & 13.19 \\

TrajCrafter~\cite{yu2025trajectorycrafter} 
& 13.49 & 0.306  & 0.428
& 53.45  & 0.0042 & 703.82 & 13.93 \\

CogNVS~\cite{chen2025cognvs}      
& 12.70 & 0.268 & 0.467 
& 81.09 & 0.0166 & 882.71 & 13.04 \\

VACE~\cite{jiang2025vace}        
& 13.26 & 0.302 & 0.476
& 92.48 & 0.0209 & 781.41 & 13.77 \\

\midrule

Lingbot-World~\cite{team2026lingbotworld} 
& 12.87 & 0.246  & 0.457
& \textbf{40.26} & \textbf{0.0013} & 742.09 & 13.07 \\

VMem~\cite{li2025vmem}          
& 12.02 & 0.232 & 0.563 
& 80.76 & 0.0138 & 867.67 & 12.13 \\

DFoT~\cite{song2025dfot}         
& 9.19 & 0.143 & 0.603 
& 82.44 & 0.0232 & 931.15 & 9.23 \\

WorldWarp~\cite{kong2025worldwarp}     
& 12.63 & 0.265 & 0.540
& 74.81 & 0.0072 & 1002.4 & 12.73  \\

\midrule

\textbf{WanAR + FixAR} 
& \textbf{13.84} & \textbf{0.320} & \textbf{0.421} 
& 47.75 & 0.0034 & \textbf{640.64} & \textbf{14.05} \\

\bottomrule
\end{tabular}}
\vspace{0.5em}
\caption{\textbf{Comparison to state-of-the-art on DL3DV~\cite{ling2024dl3dv}.} We compare our method with prior art on the DL3DV dataset which has more challenging camera motion, like sudden translation changes, and 360$^\circ$ camera pan. Despite this, our method outperforms baselines by a considerable margin. Once again, mPSNR demonstrates better camera control with our warp-and-inpaint strategy for long video generation.} 
\label{tab:sota_dl3dv}
\end{table}

\noindent\textbf{Datasets.}
We train and evaluate our model on two real-world datasets, RealEstate10K~\cite{zhou2018re10k} and DL3DV~\cite{ling2024dl3dv}, which cover diverse indoor and outdoor scenes. RealEstate10K~\cite{zhou2018re10k} contains approximately 60K videos, with ground-truth camera annotations available for both training and testing. For DL3DV~\cite{ling2024dl3dv}, since the released version does not provide camera annotations, we use a COLMAP-preprocessed training set~\cite{wang2025cut3r} with approximately 7K videos. For the test set, we process the videos with MegaSAM~\cite{li2025megasam} to obtain the corresponding camera poses and depth maps.

\begin{figure}
    \centering
    \includegraphics[width=0.85\linewidth]{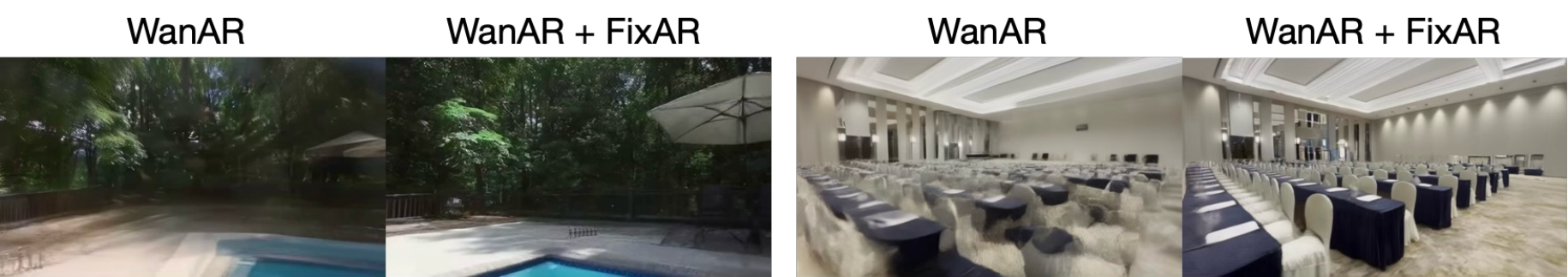}
    \caption{\textbf{Qualitative analysis of hierarchical denoising.} WanAR exhibits drift over long rollouts, while FixAR mitigates the drift and cleans up accumulated artifacts.}
    \label{fig:fixar}
\end{figure}

\begin{table*}[t]
\centering

\begin{minipage}[t]{0.47\textwidth}
\centering
\scriptsize
\setlength{\tabcolsep}{3pt}
\resizebox{\linewidth}{!}{%
\begin{tabular}{lcccccc}
\toprule
\textbf{Method} & \textbf{PSNR} $\uparrow$ & \textbf{SSIM} $\uparrow$ &
\textbf{LPIPS} $\downarrow$ & \textbf{FID} $\downarrow$ &
\textbf{KID} $\downarrow$ & \textbf{FVD} $\downarrow$ \\
\midrule
WanAR & 16.63 & 0.519 & 0.342 & 36.68 & 0.0022 & 455.87 \\
WanAR + FixAR & 16.71 & 0.521 & 0.338 & 34.55 & 0.0018 & 429.14 \\
\bottomrule
\end{tabular}%
}
\caption{\textbf{Quantitative analysis of hierarchical denoising.}
FixAR improves generation realism and leads to consistent gains across evaluation metrics.}
\label{tab:ablation1}
\end{minipage}\hfill
\begin{minipage}[t]{0.51\textwidth}
\centering
\scriptsize
\setlength{\tabcolsep}{3pt}
\resizebox{\linewidth}{!}{%
\begin{tabular}{lccccccc}
\toprule
\textbf{Method} & \textbf{PSNR} $\uparrow$ & \textbf{SSIM} $\uparrow$ &
\textbf{LPIPS} $\downarrow$ & \textbf{FID} $\downarrow$ &
\textbf{KID} $\downarrow$ & \textbf{FVD} $\downarrow$ &
\textbf{mPSNR} $\uparrow$ \\
\midrule
Warp-and-inpaint & 16.63 & 0.519 & 0.342 & 36.68 & 0.0022 & 455.87 & 18.20 \\
Plucker ray & 14.95 & 0.438 & 0.416 & 35.62 & 0.0021 & 505.49 & 15.33 \\
\bottomrule
\end{tabular}%
}
\caption{\textbf{Comparison of different conditioning strategies.}
The Plucker baseline adds a convolutional adapter to the Wan2.1-14B backbone, and is finetuned on Re10K using the same setup.}
\label{tab:pose_ablation}
\end{minipage}

\end{table*}

\noindent\textbf{Baselines.}
We compare our method with baselines from both camera-conditioned long video generation and novel-view synthesis. For camera-conditioned video generation, we compare with VMem~\cite{li2025vmem}, DFoT~\cite{song2025dfot}, Lingbot-World~\cite{team2026lingbotworld}, and WorldWarp~\cite{kong2025worldwarp}. VMem, DFoT, and Lingbot-World use Pl\"ucker rays or ray maps as camera-control representations, while WorldWarp uses 3D Gaussian rendering as camera guidance. Lingbot-World is an open-source industrial model, and we use its flagship bidirectional model for evaluation.

For novel-view synthesis, we adopt video diffusion-based methods that also use warp-and-inpaint-style camera control, including ViewCrafter~\cite{yu2025viewcrafter}, TrajCrafter~\cite{yu2025trajectorycrafter}, and CogNVS~\cite{chen2025cognvs}. We further include VACE~\cite{jiang2025vace}, a Wan-14B-based video inpainter, as an additional baseline. Since these methods are designed for fixed-length rendering, we evaluate them in an image-to-video autoregressive manner, following the same rollout setting as our method.

\noindent\textbf{Metrics.}
We follow standard evaluation protocols used in related work~\cite{xiao2025worldmem, kong2025worldwarp, li2025vmem, shen2026lyra2}, including pairwise reconstruction metrics PSNR, SSIM, and LPIPS, as well as distribution-level metrics FID and KID. We also report FVD to evaluate video generation quality. In addition, we adopt masked PSNR (mPSNR)~\cite{gao2022dycheck}, which computes PSNR only over the covisible region from the first frame. This metric better reflects camera-control accuracy without scale ambiguity. 

Since some baselines~\cite{song2025dfot, li2025vmem} only support a resolution of $256 \times 256$ and a temporal frame skip of 10, we crop and resize the outputs of all methods, including ours, to the same spatial and temporal resolution for fair comparison.

\noindent\textbf{Implementation details.}
We initialize our models from the official Wan-I2V-14B-480P checkpoint~\cite{wan2025wan} and fine-tune them using LoRA with rank 32. We use a batch size of 8 and a learning rate of $1\times10^{-4}$. For each model, including WanAR, FixAR, and InterpolatAR, we fine-tune for 10K steps on each dataset using 8 H100 GPUs at a resolution of $256 \times 256$. Training each model takes approximately one day. Further inference details are provided in Appendix~\ref{supp:Implementation}.

\vspace{-5pt}
\subsection{Comparison to state-of-the-art}
\noindent\textbf{RealEstate10K.}
Tab.~\ref{tab:sota_re10k} presents a detailed comparison between existing methods and ours on RealEstate10K. Overall, NVS methods achieve stronger reconstruction metrics, including PSNR, SSIM, LPIPS, and mPSNR, mainly because their warp-and-inpaint formulation provides more accurate camera control. In contrast, camera-conditioned long video generation methods perform better on visual fidelity metrics, such as FID, KID, and FVD, since they are typically trained with dedicated anti-drift strategies and are therefore better at preserving long-term visual quality. The importance of accurate camera control is also illustrated in Fig.~\ref{fig:qual_comp}. For example, Lingbot-World, despite being an industrial-level bidirectional video diffusion model capable of generating highly photorealistic videos, still obtains relatively poor PSNR due to inaccurate camera control. This suggests that Pl\"ucker-ray-based conditioning is not sufficiently accurate or generalizable for this setting. In contrast, our method achieves state-of-the-art performance across major metrics by combining accurate 3D-informed warp-and-inpaint camera control with hierarchical denoising for drift mitigation.

\noindent\textbf{DL3DV} is a more challenging dataset, containing fast viewpoint transitions, extreme camera trajectories, and diverse scene content. As shown in Fig.~\ref{fig:qual_comp}, many NVS methods suffer from distorted content, largely because depth estimation becomes unreliable on drifted generated frames during autoregressive rollout. In contrast, our method uses FixAR to suppress drift accumulation, ensuring that the corrected frames remain meaningful inputs for subsequent depth estimation and warping. As reported in Tab.~\ref{tab:sota_dl3dv}, our method again achieves state-of-the-art overall performance, despite marginally lower FID and KID compared with Lingbot-World. These results demonstrate that our method generalizes well to diverse indoor and outdoor scenarios with challenging camera motion.

\subsection{Ablations and analysis}

\noindent\textbf{Effect of hierarchical denoising.}
We conduct ablation studies on a 50-sequence subset of RealEstate-10K. Fig.~\ref{fig:fixar} and Tab.~\ref{tab:ablation1} show the qualitative and quantitative effects of the hierarchical denoiser. By mitigating drift, the denoiser improves generation realism and consistently improves the evaluation metrics.

\noindent\textbf{Ablations on camera pose conditioning.} We compare WanAR’s warp-and-inpaint conditioning with Plucker ray conditioning using the same backbone and training setup. As shown in Tab.~\ref{tab:pose_ablation}, warp-and-inpaint improves reconstruction quality, camera-control accuracy, and video quality, while achieving comparable FID and KID. This suggests that warp-and-inpaint provides more precise camera control in our setting. Additional ablations are reported in Appendix~\ref{supp:more_ab}.

\begin{figure}
    \centering
    \includegraphics[width=\linewidth]{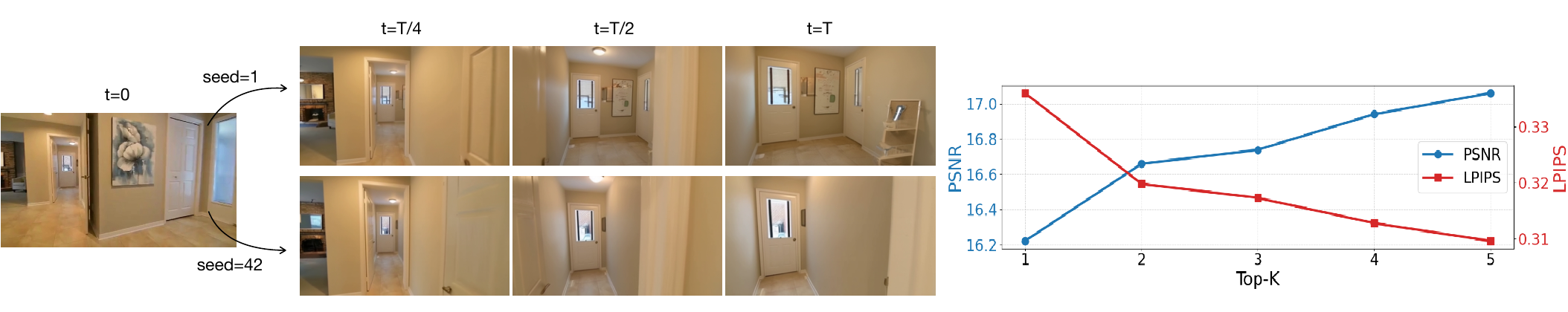}
    \caption{\textbf{Top-k evaluation.} We note that generative methods for camera-controlled video generation can produce multi-modal outputs for unseen regions of the scene as camera pans in new direction. We conduct such an evaluation for our method. Our finding is that the performance of our final predictions improves with respect to the ground-truth as more samples are drawn from WanAR (thereby covering the generation of multiple plausible 3D scenes in the future).}
    \label{fig:multimodal}
\end{figure}

\noindent\textbf{Evaluation of multi-modal generation.}
A key advantage of video diffusion models is their multi-modal generation capability, which we also observe in WanAR + FixAR. As shown in Fig.~\ref{fig:multimodal}, the same corridor scene can be plausibly generated as either a narrow straight pathway or a larger room space connected to another room by an additional door. Both outcomes are reasonable under the given input. We therefore evaluate Top-$K$ generation metrics by sampling multiple outputs with different random seeds.

\section{Discussion}
We present a simple framework for long-video, camera-controlled generation that combines autoregressive generation with hierarchical denoising. Our results show that reformulating generation as a warping-and-inpainting problem helps maintain consistency with input camera trajectories, particularly in the presence of up-to-scale ambiguity, while hierarchical denoising improves temporal stability by allowing each frame to incorporate both short- and long-range temporal information. Compared to prior approaches that rely on self-forcing or explicit pose conditioning alone, our method achieves stronger long-horizon consistency. These findings suggest that both scale consistency and long-range temporal conditioning are important for generating stable, camera-controlled videos over extended sequences.

\subsection*{Limitations} We note that our proposed warp-and-inpaint hierarchical video generation can only be used for static scenes. WanAR relies on the quality of underlying geometry, and errors in depth or reconstruction can propagate through the warping process, leading to artifacts or inconsistencies. WanAR rollouts only warp the last frame of the previous rollout and therefore do not have any memory mechanism baked in. This means that loop closure in long sequences may not be accurate. While hierarchical denoising with FixAR reduces drift, it does not fully eliminate it as our second stage operates over a finite horizon of frames too. Finally, our models trained on one dataset cannot be used for another dataset if the scale distributions between the two datasets are drastically different. We cover a detailed inference cost of our method with respect to other baselines in appendix.

\subsection*{Broader impacts}  Efficient and temporally consistent camera-controlled video synthesis could broaden access to tools for creative media production, 3D exploration in virtual reality applications, and embodied AI research, though care should be taken to mitigate risks associated with realistic synthetic video generation.





\bibliography{iclr2027_conference}
\bibliographystyle{iclr2027_conference}

\clearpage

\section*{Appendix}
\appendix
In this supplement, we include further implementation details of implementation (Section \ref{supp:Implementation}), computational cost (\ref{supp:computation_cost}), and more ablations~\ref{supp:more_ab} of our WanAR + FixAR approach. We further present a nuanced discussion of camera control in video generation (Section \ref{supp:discussion}).

\subsection{More implementation details}
\label{supp:Implementation} Given one initial image and a sequence of $L$ target cameras, our objective is to generate an $L$-frame video following the target camera trajectory. WanAR, FixAR, and InterpolatAR all use the same Wan2.1-14B architecture, but have separately trained checkpoints for their respective tasks. Unless otherwise specified, each stage operates at a resolution of $336 \times 592$ with a temporal window of $T$=49 frames.

For InterpolatAR, we place the corrected FixAR keyframes at their corresponding temporal locations and replace the missing intermediate frames with blank frames. The first frame of each input window is always an observed keyframe. InterpolatAR then predicts the missing frames conditioned on the visible keyframes (i.e. if FixAR was applied at a keyframe interval of 4 then frames 1, 5, 9 … are input to InterpolatAR which interpolates to get frames 2, 3, 4, 6, 7, 8 …).

For RealEstate10K, we generate ($L$=280) frames at 10 FPS: 1) WanAR: We subsample the target trajectory using an interval of 2 to generate the sequence with 3 autoregressive diffusion calls. 2) FixAR: We select 49 generated keyframes with an average temporal interval of approximately 6 and jointly refine them using 1 diffusion call. 3) InterpolatAR: We use an interval of 1 to reconstruct the complete 280-frame sequence, requiring 6 diffusion calls. The complete RealEstate10K pipeline uses 10 diffusion calls: 3 WanAR, 1 FixAR, and 6 InterpolatAR calls.

For DL3DV, we generate ($L$=600) frames at 30 FPS: 1) WanAR and FixAR: We subsample the target trajectory using an interval of 3 to generate the first 10s of the sequence with 2 WanAR diffusion calls. We then select 49 keyframes with an average temporal interval of 6 and jointly refine them using 1 FixAR diffusion call. Starting from the refined result, we repeat the same process with another 2 WanAR calls followed by 1 FixAR call to generate the remaining sequence. 2) InterpolatAR: We use an interval of 1 to reconstruct the complete 600-frame sequence, requiring 12 diffusion calls. The complete DL3DV pipeline uses 18 diffusion calls: 4 WanAR, 2 FixAR, and 12 InterpolatAR calls.

\subsection{Analysis of computational cost}
\label{supp:computation_cost}
We compare the computational cost of our method with the baselines. Specifically, we run inference on 280-frame RealEstate10K videos using one 80GB H100 GPU and report the average runtime, peak GPU memory, number of diffusion calls, and denoising steps per call. As shown in Tab.~\ref{tab:computation_cost}, our method has moderate runtime and GPU memory usage. We additionally report our method at $480 \times 832 \times 49$ only for a resolution-matched cost comparison; all experiments in the paper use a training resolution of $336 \times 592 \times 49$.

Although our system contains three stages, all stages reuse the same Wan2.1-14B architecture and the same input shape of $336 \times 592 \times 49$. We sequentially reload the corresponding LoRA checkpoint for each stage without introducing additional architectural modules. This unified design reduces engineering complexity. Moreover, our hierarchical design applies FixAR at a low frequency before final interpolation, limiting the total number of diffusion calls.

\begin{table}[t]
\centering
\scriptsize
\setlength{\tabcolsep}{3pt}
\resizebox{0.95\textwidth}{!}{
\begin{tabular}{lccccc}
\toprule
\textbf{Method} & \textbf{Backbone} & \textbf{Resolution} & \textbf{GPU Memory} & \textbf{Run Time} & \textbf{\# Diffusion Calls} \\
\midrule

ViewCrafter & StableVideoDiffusion-2B & $576\times1024\times25$ & 22GB & 15min & 12 \\
TrajCrafter & CogVideoX-5B & $480\times720\times49$ & 48GB & 18min & 6 \\
CogNVS & CogVideoX-5B & $480\times720\times49$ & 23GB & 16min & 6 \\
VACE & Wan2.1-14B & $480\times832\times49$ & 51GB & 19min & 6 \\
VMem & SEVA-1.3B & $576\times576\times8$ & 6GB & 4min & 6 \\
DFoT & DiT-XL/2-0.68B & $256\times256\times8$ & 7GB & 5min & 44 \\
WorldWarp & Wan2.1-1.3B & $480\times832\times81$ & 36GB & 8min & 4 \\
Lingbot-World & Wan2.2-14B & $480\times832\times281$ & 75GB & 38min & 1 \\
Ours & Wan2.1-14B & $336\times592\times49$ & 41GB & 12min & 10 \\
& Wan2.1-14B & $480\times832\times49$ & 48GB & 17min & 10 \\
\bottomrule
\end{tabular}}
\vspace{0.5em}
\caption{\textbf{Computational cost comparison with baselines.} Runtime and peak GPU memory are measured on 280-frame RealEstate10K videos using a single 80GB H100 GPU. We also report our method at $480\times832\times49$ for a resolution-matched comparison.}
\label{tab:computation_cost}
\end{table}

\begin{table}[t]
\centering
\scriptsize
\setlength{\tabcolsep}{3pt}
\resizebox{0.95\textwidth}{!}{
\begin{tabular}{lccccccc}
\toprule
\textbf{Setting} & \textbf{Keyframe Interval} & \textbf{PSNR} $\uparrow$ & \textbf{SSIM} $\uparrow$ & \textbf{LPIPS} $\downarrow$ & \textbf{FID} $\downarrow$ & \textbf{KID} $\downarrow$ & \textbf{FVD} $\downarrow$ \\
\midrule

49-length FixAR & 2 & 16.38 & 0.508 & 0.336 & 34.33 & 0.0021 & 424.94 \\
& 4 & 16.78 & 0.522 & 0.333 & 34.06 & 0.0017 & 428.02 \\
& 6 & 16.71 & 0.521 & 0.338 & 34.55 & 0.0018 & 429.14 \\

25-length FixAR & 4 & 16.11 & 0.490 & 0.340 & 38.02 & 0.0021 & 461.49 \\
& 8 & 15.91 & 0.487 & 0.351 & 38.72 & 0.0022 & 488.46 \\
& 12 & 15.46 & 0.471 & 0.384 & 40.07 & 0.0030 & 561.53 \\

\bottomrule
\end{tabular}}
\vspace{0.5em}
\caption{\textbf{Ablation on FixAR keyframe interval and sequence length.} We compare 49-frame and 25-frame FixAR variants under different keyframe intervals. The 49-frame model remains stable across its training range, while larger intervals degrade the performance of the 25-frame model.}
\label{tab:fixar_interval}
\end{table}

\begin{table}[t]
\centering
\scriptsize
\setlength{\tabcolsep}{3pt}
\resizebox{0.75\textwidth}{!}{
\begin{tabular}{lcccccc}
\toprule
\textbf{Method} & \textbf{PSNR} $\uparrow$ & \textbf{SSIM} $\uparrow$ & \textbf{LPIPS} $\downarrow$ & \textbf{FID} $\downarrow$ & \textbf{KID} $\downarrow$ & \textbf{FVD} $\downarrow$ \\
\midrule

MoGe2 & 20.22 & 0.654 & 0.177 & 21.90 & 0.0013 & 130.58 \\
Depth Anything v3 & 20.46 & 0.666 & 0.172 & 21.77 & 0.0014 & 129.02 \\
MegaSAM & 20.74 & 0.679 & 0.170 & 21.92 & 0.0015 & 125.77 \\

\bottomrule
\end{tabular}}
\vspace{0.5em}
\caption{\textbf{Comparison of different geometry estimation methods.} We initialize WanAR using the first-frame depth estimated by each method and report performance after one rollout. Results are similar across estimators, indicating limited sensitivity to the choice of depth estimator.}
\label{tab:depth_ablation}
\end{table}

\subsection{Additional ablations}
\label{supp:more_ab}

\paragraph{Sensitivity of FixAR's keyframe interval.} Our default FixAR processes 49 frames, is trained with keyframe intervals ranging from 2 to 6, and uses an interval of 6 during inference. We additionally train a 25-frame FixAR with keyframe intervals ranging from 4 to 12 to study the effects of input length and larger keyframe intervals. As shown in Tab.~\ref{tab:fixar_interval}, the 49-frame FixAR performs consistently across intervals from 2 to 6, indicating that it is not sensitive to the precise interval within its training range. In contrast, the performance of the 25-frame FixAR decreases when the interval increases from 4 to 8 or 12. With very sparse keyframes, the temporal correspondence between adjacent input frames becomes weaker, making refinement more difficult. The 49-frame FixAR also consistently outperforms the 25-frame variant; a longer input sequence can cover a longer temporal range while retaining relatively dense keyframes. 

\paragraph{Effect of different depth estimators.} We estimate the depth of the first frame using each method and then perform one WanAR rollout. As shown in Tab.~\ref{tab:depth_ablation}, the performance differences among the three depth estimators is relatively small, indicating that WanAR is robust to the choice of depth estimator – it can recover from errors introduced by depth estimators. Depth Anything V3 and MegaSAM depth provide slightly better reconstruction and video-quality metrics.

\subsection{Extended discussion on camera-controlled video generation}
\label{supp:discussion}

In this section, we elucidate our claim that camera-controlled video generation is a special case of novel-view synthesis (NVS) and discuss relevant paradigms for evaluation which are inspired by the literature in NVS.

\subsubsection*{Case I: Image-to-Video generation with pose / geometry information}

When input observations are accompanied by reconstruction information, such as calibrated camera poses and/or 3D geometry defined in a fixed coordinate system, future camera trajectories should also be specified in that same coordinate system. This corresponds to the standard NeRF-style setting, where methods are conditioned on input images together with calibrated camera poses.
More generally, this setup may also be instantiated using a single image paired with depth or XYZ information. For example, given an estimate of scene geometry, future camera motion can be defined relative to the reconstructed scene structure (e.g., moving a fixed distance toward an observed object). Framing camera control in this coordinate-aware manner removes the need to reference future ground-truth geometry during alignment and provides a cleaner formulation of camera-conditioned video generation.

\subsubsection*{Case II: Image-to-Video generation without pose information}
\begin{figure}[h!]
    \centering
    \includegraphics[width=0.55\linewidth]{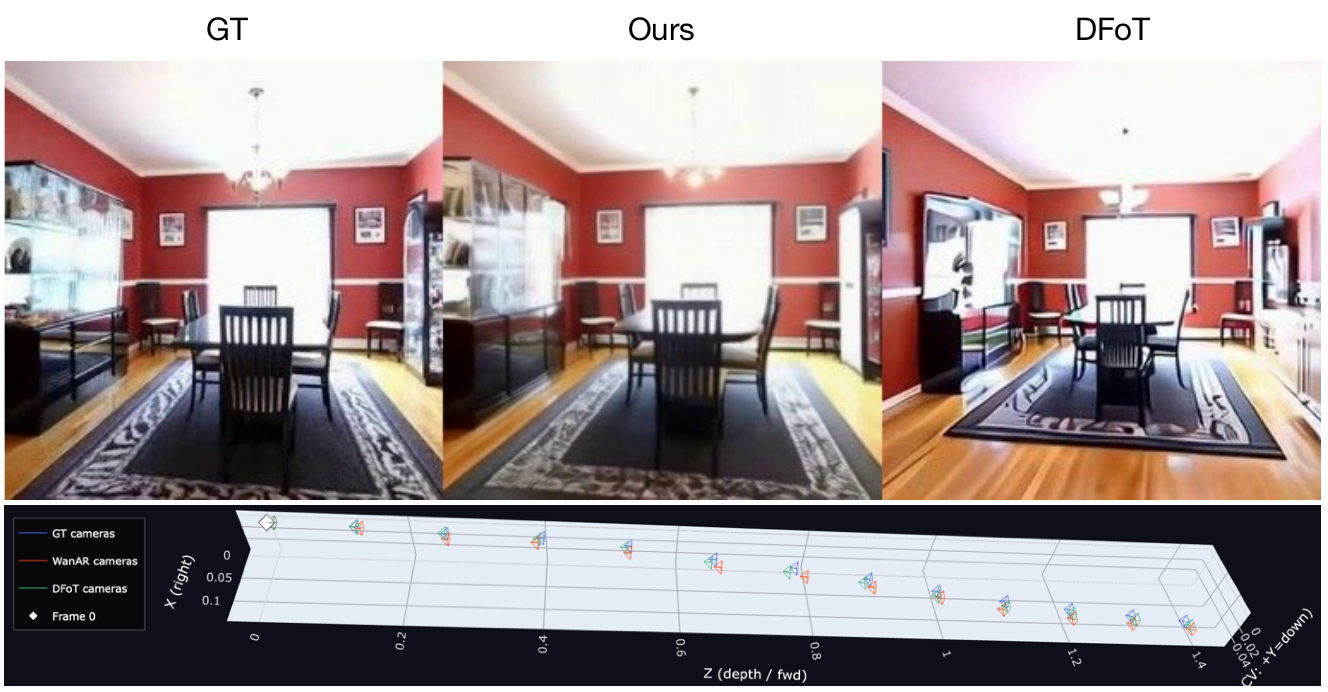}
    \caption{We identify a key limitation of existing camera-conditioned methods. When explicit 6-DOF poses are used as conditioning—e.g., via Plücker coordinates or ray maps \cite{song2025dfot, team2026lingbotworld}—it is difficult to enforce a consistent scale between the scene and the camera trajectory. As a result, generated videos may exhibit visual inconsistencies (\textbf{top}), reflecting scenes rendered under an arbitrary pose scale. Importantly, standard camera-control metrics \cite{xiao2025worldmem} perform a \textit{scale alignment} with ground-truth camera poses prior to evaluation. Consequently, even when predicted camera trajectories are incorrect up to a global scale factor, they can still achieve low error (\textbf{bottom}), masking this failure mode. Current methods evaluating performance with the PSNR metric circumvent this issue, as they overfit to the pose distribution in a particular datasets, which indicates their limited generalization to other scene or camera trajectories. In our approach, we mitigate this issue by aligning MoGE depth predictions to ground-truth depth for each rollout, and by adopting a warping-and-inpainting formulation for long-video generation. We also adopt novel-view synthesis protocols for evaluation, and instead benchmark methods for camera-control with the masked PSNR (mPSNR) metric.}
    \label{fig:scaleinconsistency}
\end{figure}

In contrast, when only input images are provided without associated poses or geometric information of the input (such as the current setup of camera-controlled video generation works including ours), the problem more closely resembles unposed novel-view synthesis. In this setting, one possible approach is to require future camera trajectories to be specified in a metric coordinate system to avoid key pitfalls of prior art (c.f. Figure \ref{fig:scaleinconsistency}).
An alternative is to follow protocols of recent benchmarks \cite{wriva1} under whose formulation:
(i) methods reconstruct the input observations within a locally consistent coordinate system;
(ii) the benchmark provides a similarity transform aligning the reconstruction to the ground-truth frame; and
(iii) future camera queries are specified within the local reconstructed coordinate system.
Our current training, inference and benchmarking protocol is based on this formulation.

\end{document}